\documentclass[runningheads]{llncs}
\usepackage[T1]{fontenc}
\usepackage{graphicx}
\usepackage{tabularx}
\usepackage{hyperref}
\newcolumntype{Y}{>{\centering\arraybackslash}X}

\usepackage{booktabs,multirow,siunitx}
\newcolumntype{d}{S[table-format=1.2]}

\usepackage{etoolbox}
\makeatletter
\patchcmd{\thebibliography}
  {\leftmargin\labelwidth}
  {\leftmargin=0pt}{}{}
\patchcmd{\thebibliography}
  {\advance\leftmargin\labelsep}
  {}{}{}
\makeatother

\usepackage{booktabs} 
\usepackage{tcolorbox} 
\usepackage{graphicx} 
\usepackage{amsmath} 
\usepackage{amssymb} 
\usepackage{xcolor} 
\usepackage{multirow} 
\usepackage{array} 

\tcbset{colback=gray!5, colframe=black!50, boxrule=0.5pt, arc=2pt, left=2pt, right=2pt, top=2pt, bottom=2pt}
\begin{document}
\title{Towards Reliable Machine Translation: Scaling LLMs for Critical Error Detection and Safety}
\titlerunning{Towards Reliable Machine Translation}
%
\newcommand{\equalcontrib}{\textsuperscript{*}}

\author{
  Muskaan Chopra\inst{1,3}\equalcontrib \and
  Lorenz Sparrenberg\inst{1,3}\equalcontrib \and
  Rafet Sifa\inst{1,2,3}
}
\authorrunning{M. Chopra et al.}

\institute{
  Rheinische Friedrich-Wilhelms-Universität Bonn, Bonn, Germany
  \email{chopra.muskaan@uni-bonn.de}\\
  \and
  Fraunhofer IAIS, Sankt Augustin, Germany \\
  \and
  Lamarr Institute for Machine Learning and Artificial Intelligence, Germany
}

\renewcommand\thefootnote{*}
\footnotetext{Equal contribution.}
\maketitle      
\begin{abstract}
Machine Translation (MT) plays a pivotal role in cross-lingual information access, public policy communication, and equitable knowledge dissemination. However, critical meaning errors, such as factual distortions, intent reversals, or biased translations, can undermine the reliability, fairness, and safety of multilingual systems. In this work, we explore the capacity of instruction-tuned Large Language Models (LLMs) to detect such critical errors, evaluating models across a range of parameters using the publicly accessible data sets. Our findings show that model scaling and adaptation strategies (zero-shot, few-shot, fine-tuning) yield consistent improvements, outperforming encoder-only baselines like XLM-R and ModernBERT. We argue that improving critical error detection in MT contributes to safer, more trustworthy, and socially accountable information systems by reducing the risk of disinformation, miscommunication, and linguistic harm, especially in high-stakes or underrepresented contexts. This work positions error detection not merely as a technical challenge, but as a necessary safeguard in the pursuit of just and responsible multilingual AI. The code will be made available at GitHub.\footnote{\href{https://github.com/muskaan712/ecir26-ced}{https://github.com/muskaan712/ecir26-ced}}

\end{abstract}
\section{Introduction}

Large Language Models (LLMs) and neural Machine Translation (MT) systems are increasingly woven into everyday information access, from wearable devices such as Apple AirPods Pro 3~\cite{airpods3} and Ray-Ban Meta Glasses~\cite{metaglasses} to multilingual assistants and professional communication platforms. As these technologies mediate global communication, their reliability becomes central to public trust, accessibility, and equitable participation in the digital sphere. Yet even advanced MT systems continue to produce \textit{critical meaning errors}, mistranslations that invert intent, distort factual content, or embed social bias. Such errors can be especially harmful in domains like health \cite{health}, legal aid \cite{legal}, and finance \cite{finance} response, where users may depend on automated translation for essential information.

Traditional metrics such as BLEU~\cite{papineni2002bleu}, METEOR~\cite{banerjee2005meteor}, and COMET~\cite{rei2020comet} offer coarse measures of translation quality but fail to capture meaning-critical distortions~\cite{lommel2014multidimensional}. This gap motivates research on \textit{Critical Error Detection (CED)}, identifying translations that introduce semantically or socially consequential deviations. Earlier work on contradiction detection in German~\cite{9003090,pielka2020contradiction} established its technical and societal importance, while recent LLM-based verification approaches~\cite{pielka2025translation} highlight new opportunities for scalable and context-aware detection.

However, most existing CED systems rely on encoder-only models such as XLM-R~\cite{conneau2020unsupervised} or ModernBERT~\cite{modernbert}, which struggle to generalize across domains and long contexts. In contrast, instruction-tuned LLMs demonstrate strong reasoning and factual alignment abilities. In this study, we systematically examine how model scale and adaptation regime (zero-shot, few-shot, fine-tuned) affect CED performance across WMT 21/22 benchmarks~\cite{federmann2021findings,federmann2022findings} and SynCED-En De 2025 \cite{anonymous}. 

Our results show that scaling and instruction alignment yield consistent improvements, surpassing encoder-only baselines. We frame this progress not merely as a technical advance but as a contribution toward \textit{trustworthy multilingual information systems} that support fairness, inclusivity, and civic safety in global communication. \textbf{Our contributions are threefold:}
\begin{enumerate}
    \item We present the first cross-model scaling study of instruction-tuned LLMs for Critical Error Detection in Machine Translation.
    \item We empirically compare zero-shot, few-shot, and fine-tuned regimes across multiple model families and translation benchmarks.
    \item We discuss the broader societal implications of reliable error detection as a foundation for equitable and accountable multilingual AI.
\end{enumerate}

\subsection*{Theory of Change}

Critical meaning preservation is essential for fairness and trust in multilingual communication. Undetected translation errors can distort public messages, misinform users, or marginalize communities. Our work proposes instruction-aligned LLMs as practical safeguards that flag high-risk translations in real time, enabling journalists, NGOs, and institutions to validate cross-lingual content before dissemination.

For impact to materialize, translation workflows must retain human oversight and ensure interpretability of model outputs. While automation bias and false positives remain risks, calibrated scoring and human-AI review loops can mitigate them. By integrating CED into the translation pipeline, we aim to strengthen the integrity of digital communication and advance the Information Retrieval-for-Good vision of information systems that are transparent, reliable, and socially responsible.

\section{Related Work}

\subsection{Critical Error Detection and Quality Estimation}

Quality Estimation (QE) aims to predict translation quality without human references, forming the foundation for automatic MT reliability assessment.  
Early shared tasks at WMT~\cite{specia2020findings,federmann2021findings,federmann2022findings} established multilingual benchmarks and metrics such as BLEU~\cite{papineni2002bleu}, METEOR~\cite{banerjee2005meteor}, and COMET~\cite{rei2020comet}.  
Although these metrics correlate with adequacy and fluency, they struggle to capture \textit{meaning-critical distortions} that alter factual or pragmatic content~\cite{lommel2014multidimensional}.  

Recent efforts have extended QE toward \textit{Critical Error Detection (CED)} by explicitly identifying semantic and factual inconsistencies.  
Early contradiction-detection studies in German~\cite{9003090,pielka2020contradiction} demonstrated that translation-driven approaches can expose cross-lingual inconsistencies often missed by surface metrics.  
Later work introduced domain-specific validation of financial translations using large language models~\cite{pielka2025translation} and informed pre-training for CED~\cite{pucknat2022informed}. 
Collectively, these studies emphasize the need for contextual reasoning and semantic alignment beyond token-level similarity.

\subsection{Instruction-Tuned and Scaled LLMs for Translation Evaluation}

Transformer-based encoders such as XLM-R~\cite{conneau2020unsupervised} and ModernBERT~\cite{modernbert} advanced multilingual representation learning and long-context modeling but still face limitations in fine-grained semantic reasoning.  
Recent advances show that instruction-tuned LLMs can outperform traditional QE models both as evaluators~\cite{kocmi2023lm_eval,lu2023erroranalysis} and as detectors of translation-level failures.  
Work such as Peng et al.~\cite{peng2023towards} highlights that ChatGPT-style models capture discourse-level dependencies absent in encoder-only architectures.  
These findings motivate scaling experiments that examine how model size and instruction alignment influence robustness, generalization, and efficiency in CED.

\subsection{Trustworthy and Socially Responsible MT}

Beyond accuracy, MT evaluation increasingly intersects with fairness, transparency, and accountability.  
Prior research has examined human-in-the-loop evaluation~\cite{graham2017can}, explainable CED datasets~\cite{jung2024explainable}, and multilingual QE resources for low-resource or safety-critical contexts~\cite{fomicheva2020mlqe}.  
Our work builds on this trajectory by situating CED within the \textit{IR-for-Good} paradigm, improving translation reliability not merely as a technical goal but as an enabler of equitable information access and civic inclusion.  
We extend prior literature by quantifying how instruction-tuned LLMs scale toward more socially accountable MT error detection.

\section{Methodology}

\begin{figure}[htbp]
    \centering
    \includegraphics[width=1\linewidth]{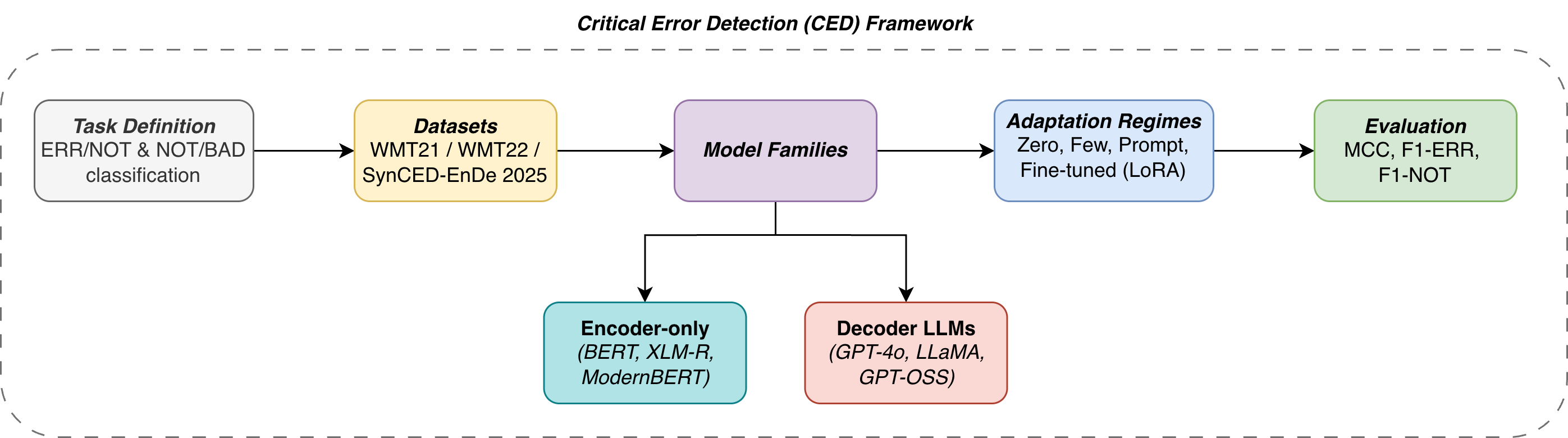}
    \caption{Conceptual overview of the proposed Critical Error Detection (CED) framework.
    The pipeline progresses from task definition and dataset sources to diverse model families, encoder-only and decoder LLMs, followed by adaptation regimes and evaluation.
    Together, these components support safer multilingual information access through meaning-preserving translation.}
    \label{fig:ced_framework}
\end{figure}

As illustrated in Figure~\ref{fig:ced_framework}, the proposed Critical Error Detection (CED) framework comprises task definition, model families (including encoder-only and decoder LLMs), adaptation regimes, and evaluation metrics.

\subsection{Task Definition}

We formulate \textit{Critical Error Detection (CED)} in Machine Translation (MT) as a binary classification task: given a source sentence $S$ and its translation $T$, a model must determine whether $T$ introduces a meaning-critical error (\texttt{ERR/BAD}) or preserves the intended meaning (\texttt{NOT/OK}). Unlike traditional Quality Estimation frameworks~\cite{specia2020findings,federmann2021findings}, which assign continuous adequacy scores, CED explicitly targets semantic distortions that may alter intent, factuality, or bias. This framing extends prior research in contradiction detection for German~\cite{9003090,pielka2020contradiction} and domain-specific translation verification~\cite{pielka2025translation}, situating CED within the broader goals of safe and socially responsible multilingual information access.

\subsection{Model Families}

We evaluate two broad classes of architectures and cover specific releases used in our study. 
\textit{Encoder-only baselines} comprise BERT-base~\cite{bert}, ModernBERT-base and -large~\cite{modernbert}, mmBERT~\cite{mmbert}, and XLM-R-large~\cite{conneau2020unsupervised}. 
These models offer strong multilingual sentence representations but limited generative reasoning. 
\textit{Decoder LLMs} include GPT-4o and GPT-4o-mini~\cite{openai2024gpt4,openai2024gpt4o}, LLaMA-3.1-8B Instruct and LLaMA-3.3-70B Instruct (GGUF)~\cite{meta2024llama3,meta2025llama33}, and GPT-OSS-20B / GPT-OSS-120B~\cite{gptoss2025report,gptoss2025lora}. 
Across regimes (zero-shot, few-shot, prompt-engineered, and voting committees), we run experiments on \emph{all} of the above models. 
For parameter-efficient fine-tuning, we focus on three representative LLMs that span model families and parameters: GPT-4o-mini, LLaMA-3.1-8B Instruct, and GPT-OSS-20B.

\subsection{Adaptation Regimes}

We examine four adaptation regimes that progressively increase task supervision. In the \textit{zero-shot} setting, a concise instruction prompts the model to classify each translation as \texttt{ERR/BAD} or \texttt{NOT/OK}. The \textit{few-shot} setting augments this with five to eight balanced examples to probe in-context learning. A \textit{prompt-engineered} variant enriches supervision through domain-specific cues emphasizing factual consistency, polarity preservation, and bias sensitivity. Finally, a \textit{fine-tuned} regime applies parameter-efficient LoRA adaptation on approximately 170,000 translation pairs (8000 WMT-21, 155,511 WMT-22, and 8000 SynCED-EnDe 2025), implemented via the Unsloth framework~\cite{unsloth2024}. Fine-tuned models are evaluated under zero-shot-style instructions to isolate the effects of parameter updating from in-context reasoning. For robustness, majority voting (committee size = 3) is applied across LLMs in zero- and few-shot settings, and across multiple fine-tuned runs for stability evaluation.

\subsection{Evaluation Metrics}

Performance is primarily measured via the Matthews Correlation Coefficient (MCC), which remains stable under class imbalance. Complementary F1-scores for both \texttt{ERR/BAD} and \texttt{NOT/OK} classes follow standard MT evaluation practices~\cite{lommel2014multidimensional,pucknat2022informed}. Metrics are computed on disjoint test splits to ensure comparability across datasets and model scales.

\section{Datasets}

We evaluate our models on three English-German datasets that together balance scale, annotation reliability, and linguistic coverage: WMT~21 \cite{wmt21_1,wmt21_2}, WMT~22 \cite{wmt22}, and SynCED  EnDe~2025~\cite{anonymous}. Each dataset provides parallel source-translation pairs labeled for meaning-critical errors, where \texttt{ERR/BAD} marks semantic distortions and \texttt{NOT/OK} indicates faithful translations.

\textbf{WMT-22}\cite{wmt22} expanded the task to 155{,}511 training and 17{,}280 development pairs. The training set comprises 146{,}574 \texttt{OK} and 8{,}937 \texttt{BAD} samples, while development includes 16{,}329 \texttt{OK} and 951 \texttt{BAD}. This large imbalance reflects real-world MT reliability scenarios.

\textbf{SynCED-EnDe~2025}\cite{anonymous} provides 8{,}000 training and 1{,}000 development pairs with balanced labels: 4{,}000 \texttt{NOT} and 4{,}000 \texttt{ERR} in training, 500 each in development. It combines synthetic perturbations with manual validation for consistent, high-quality supervision.

For fine-tuning, we merge the training samples from each dataset into a unified corpus of 171,511 examples. Evaluation always uses the original development splits to maintain strict train-test separation and preserve domain diversity.

\section{Experimental Setup}
\label{sec:exp_setup}

This work evaluates encoder-only and decoder-based LLMs under multiple adaptation regimes for Critical Error Detection (CED) in EN$\rightarrow$DE translation. Models are compared across \textbf{WMT-21}, \textbf{WMT-22}, and \textbf{SynCED-EnDe 2025} using Matthews Correlation Coefficient (MCC), class-wise F1 (\texttt{ERR/BAD}, \texttt{NOT/OK}). The evaluation investigates whether instruction alignment and model scaling improve multilingual reliability and fairness-key goals within the IR-for-Good agenda.

\subsection{Compute, Data, and Decoding}
All experiments were conducted on NVIDIA A100 and H100 nodes. Fine-tuning used approximately 170{,}000 training pairs from WMT-21, WMT-22, and SynCED-EnDe 2025. \emph{GPT-4o-mini} was fine-tuned through the OpenAI API, while \emph{LLaMA-3.1-8B Instruct} and \emph{GPT-OSS-20B} were trained locally with LoRA (Unsloth). FP16 was used only for GPT-OSS-20B. Unless specified otherwise, decoding used a fixed temperature \(t{=}0.0\); committee ensembles (size~3) used \(t{=}0.2\) to promote diversity.

\subsection{Prompt Variants}
We evaluate models under three instruction regimes of increasing contextual support. \textbf{P0} denotes zero-shot evaluation with a concise task description; \textbf{P1} augments the same instruction with eight few-shot examples (five \texttt{ERR/BAD}, three \texttt{NOT/OK}). \textbf{P2-P4} represent tuned, family-specific variants adapted to GPT, LLaMA, and GPT-OSS models, improving label alignment and mitigating instruction underfitting. Majority-voting ensembles combine three stochastic generations (\(t{=}0.2\)) by simple majority to enhance robustness.

\begin{tcolorbox}[colback=gray!5,colframe=black!50,
title=P0 / P1 : Core Instruction and Few-shot Extension\label{prompt:p0p1}]
\scriptsize
You are a precise translation evaluator.\\
Given an English sentence (EN) and its German translation (DE), respond with exactly one token:\\
\textbf{ERR} if DE has a major error (meaning shift, omission, or inaccuracy), or \textbf{NOT} if it is accurate or only has minor imperfections.\\
Do not add any explanation, punctuation, or additional text.\\[4pt]
\textbf{Few-shot variant (P1):} Identical to P0, but prepends eight labeled exemplars 
(5 \texttt{ERR}, 3 \texttt{NOT}) before the query instance to provide minimal supervision.
\end{tcolorbox}

\noindent
Higher-index prompts (\textbf{P2-P4}) are tuned, family-specific templates 
for GPT, LLaMA, and GPT-OSS models. The full texts are provided in Appendix~\ref{app:prompts}.

\subsection{Study Design}
We report results across four regimes:
(i) encoder-only baselines, (ii) preliminary decoder LLMs experiments (zero- and few-shot),
(iii) prompt tuning and committee voting, and (iv) fine-tuning with zero-shot inference.
Each table stacks all three datasets vertically for compactness and ease of comparison.

\section{Results and Discussion}
\label{sec:results}

\paragraph{Encoder-Only Baselines.}
Encoders provide strong baselines for CED, particularly on WMT-22 and SynCED-EnDe 2025, where \emph{mmBERT} and \emph{XLM-R-large} achieve MCC up to 0.88 (Table~\ref{tab:enc_only_stacked}). These results confirm that contextual representation models capture much of the signal required for translation error discrimination. 
Notably, WMT-22 scores are consistently higher than WMT-21 due to clearer lexical and adequacy cues in the \texttt{BAD/OK} annotation scheme, which favors representation-based models.  
In contrast, WMT-21 contains subtler meaning shifts, leading to lower \texttt{ERR/BAD} recall. SynCED-EnDe 2025 exhibits high and stable scores because its curated samples contain less label ambiguity and stronger contextual cues for error detection.

\begin{table}[htbp]
\centering
\caption{Encoder-only baselines (MCC, F1-ERR, F1-NOT). For WMT-22, ERR/NOT correspond to BAD/OK.}
\label{tab:enc_only_stacked}
\scriptsize
\setlength{\tabcolsep}{12pt}
\renewcommand{\arraystretch}{1.1}
\begin{tabular}{ll d d d}
\toprule
\textbf{Dataset} & \textbf{Model} &
\multicolumn{1}{c}{\textbf{MCC}} &
\multicolumn{1}{c}{\textbf{F1-ERR}} &
\multicolumn{1}{c}{\textbf{F1-NOT}} \\
\midrule

\multirow{5}{*}{WMT-21}
 & BERT-base            & 0.38 & 0.53 & 0.83 \\
 & ModernBERT-base      & 0.37 & \textbf{0.56} & 0.79 \\
 & ModernBERT-large     & 0.38 & 0.53 & 0.83 \\
 & mmBERT               & 0.38 & 0.53 & 0.82 \\
 & \textbf{XLM-R-large} & \textbf{0.46} & 0.59 & \textbf{0.85} \\
\midrule

\multirow{5}{*}{WMT-22}
 & BERT-base            & 0.69 & 0.70 & 0.98 \\
 & ModernBERT-base      & 0.84 & 0.84 & 0.99 \\
 & ModernBERT-large     & 0.86 & 0.87 & 0.99 \\
 & \textbf{mmBERT}      & \textbf{0.88} & \textbf{0.88} & \textbf{0.99} \\
 & XLM-R-large          & 0.87 & 0.88 & \textbf{0.99} \\
\midrule

\multirow{5}{*}{SynCED-EnDe 2025}
 & BERT-base            & 0.72 & 0.83 & 0.86 \\
 & ModernBERT-base      & 0.49 & 0.77 & 0.69 \\
 & ModernBERT-large     & 0.52 & 0.78 & 0.72 \\
 & mmBERT               & \textbf{0.81} & 0.88 & 0.91 \\
 & \textbf{XLM-R-large} & \textbf{0.81} & \textbf{0.89} & \textbf{0.91} \\
\bottomrule
\end{tabular}
\end{table}

\paragraph{Preliminary Decoder LLMs (Zero/Few-Shot).}
Decoder LLMs show competitive zero-shot results (P0), particularly for larger instruction-tuned models such as \emph{GPT-4o} and \emph{LLaMA-3.3-70B}. Few-shot prompts (P1) stabilize predictions and improve minority-class recall, indicating the benefits of minimal supervision (Table~\ref{tab:dec_prelim_full}).  
The improvement from P0 to P1 arises mainly from increased \texttt{ERR/BAD} F1, as the few-shot examples act as localized guidance-teaching smaller models to attend to meaning shifts, omissions, and factual reversals that they might otherwise overlook.  
Interestingly, while large models like GPT-4o remain consistent across settings, smaller ones such as LLaMA-3.1-8B benefit substantially from P1-highlighting that \textbf{alignment, not scale, drives reliability}.  
Minor drops in some cases (e.g., GPT-OSS-120B on WMT-21) stem from \textbf{instruction interference}, where long prompts overfit stylistic cues rather than semantic consistency.

\begin{table}[h]
\centering
\caption{Decoder LLMs - Zero-shot (P0) and Few-shot (P1) results (MCC, F1-ERR, F1-NOT). 
For WMT-22, ERR/NOT corresponds to BAD/OK.}
\label{tab:dec_prelim_full}
\scriptsize
\setlength{\tabcolsep}{9pt}
\renewcommand{\arraystretch}{1.1}
\begin{tabular}{l l l d d d}
\toprule
\textbf{Dataset} & \textbf{Model} & \textbf{Setting} &
\multicolumn{1}{c}{\textbf{MCC}} &
\multicolumn{1}{c}{\textbf{F1-ERR}} &
\multicolumn{1}{c}{\textbf{F1-NOT}} \\
\midrule

\multirow{12}{*}{WMT-21}
 & GPT-4o              & P0 & 0.33 & 0.55 & 0.75 \\
 & GPT-4o-mini         & P0 & 0.30 & 0.53 & 0.68 \\
 & LLaMA-3.3-70B      & P0 & 0.34 & 0.50 & \textbf{0.83} \\
 & LLaMA-3.1-8B       & P0 & 0.16 & 0.47 & 0.35 \\
 & GPT-OSS-120B       & P0 & 0.34 & \textbf{0.55} & 0.81 \\
 & GPT-OSS-20B        & P0 & 0.25 & 0.51 & 0.61 \\
 & \textbf{GPT-OSS-120B}       & P1 & \textbf{0.36} & 0.53 & 0.81 \\
 & GPT-4o              & P1 & 0.35 & 0.56 & 0.69 \\
 & LLaMA-3.3-70B      & P1 & 0.34 & 0.53 & 0.80 \\
 & LLaMA-3.1-8B       & P1 & 0.26 & 0.52 & 0.59 \\
 & GPT-4o-mini         & P1 & 0.31 & 0.54 & 0.68 \\
 & GPT-OSS-20B        & P1 & 0.33 & 0.54 & 0.76 \\
\midrule

\multirow{12}{*}{WMT-22}
 & GPT-4o              & P0 & 0.55 & 0.52 & 0.94 \\
 & GPT-4o-mini         & P0 & 0.43 & 0.39 & 0.91 \\
 & \textbf{LLaMA-3.3-70B}      & P0 & \textbf{0.62} & \textbf{0.62} & \textbf{0.96} \\
 & LLaMA-3.1-8B       & P0 & 0.11 & 0.12 & 0.32 \\
 & GPT-OSS-120B       & P0 & 0.23 & 0.22 & 0.81 \\
 & GPT-OSS-20B        & P0 & 0.24 & 0.21 & 0.75 \\
 & LLaMA-3.3-70B      & P1 & \textbf{0.62} & \textbf{0.61} & \textbf{0.97} \\
 & GPT-4o              & P1 & 0.49 & 0.44 & 0.92 \\
 & GPT-4o-mini         & P1 & 0.37 & 0.32 & 0.87 \\
 & LLaMA-3.1-8B       & P1 & 0.20 & 0.17 & 0.67 \\
 & GPT-OSS-120B       & P1 & 0.35 & 0.32 & 0.88 \\
 & GPT-OSS-20B        & P1 & 0.28 & 0.25 & 0.83 \\
\midrule

\multirow{12}{*}{SynCED-EnDe 2025}
 & GPT-4o              & P0 & 0.95 & 0.96 & 0.96 \\
 & GPT-4o-mini         & P0 & 0.92 & 0.96 & 0.96 \\
 & \textbf{LLaMA-3.3-70B}      & P0 & \textbf{0.96} & \textbf{0.98} & \textbf{0.98} \\
 & LLaMA-3.1-8B       & P0 & 0.53 & 0.78 & 0.65 \\
 & GPT-OSS-120B       & P0 & 0.84 & 0.92 & 0.92 \\
 & GPT-OSS-20B        & P0 & 0.67 & 0.84 & 0.83 \\
 & GPT-4o              & P1 & 0.95 & 0.98 & 0.98 \\
 & GPT-4o-mini         & P1 & 0.91 & 0.96 & 0.95 \\
 & LLaMA-3.3-70B      & P1 & \textbf{0.96} & \textbf{0.98} & \textbf{0.98} \\
 & LLaMA-3.1-8B       & P1 & 0.62 & 0.82 & 0.73 \\
 & GPT-OSS-120B       & P1 & 0.92 & 0.96 & 0.96 \\
 & GPT-OSS-20B        & P1 & 0.75 & 0.88 & 0.87 \\
\bottomrule
\end{tabular}
\end{table}

\paragraph{Prompt Tuning and Committee Voting.}
Prompt tuning (P2-P4) generally enhances performance across families, especially for GPT-family models and LLaMA-3.1-8B, while LLaMA-3.3-70B and GPT-OSS occasionally experience slight declines (Table~\ref{tab:tuned_comm_stacked}).  
This is expected: instruction-tuned templates improve alignment for smaller or underfit models by clarifying the task boundary (ERR vs.\ NOT), but can introduce mild over-specification for larger models that already possess strong internal alignment, leading to marginally lower MCC.  
Committee voting (size~3, \(t{=}0.2\)) mitigates this by averaging across stochastic generations-reducing variance and correcting individual flips. This ensemble effect boosts stability, particularly on noisier data (e.g., WMT-21), resulting in smoother class balance between \texttt{ERR/BAD} and \texttt{NOT/OK} predictions.

\paragraph{Fine-Tuning with Zero-Shot Inference.}
Fine-tuned models (trained on $\sim$170K pairs) outperform all baselines (Table~\ref{tab:finetuned_stacked}). \emph{GPT-4o-mini} achieves the highest consistency (MCC 0.92-0.94; F1 0.94-0.96), followed by \emph{LLaMA-3.1-8B} and \emph{GPT-OSS-20B}.  
The gains are most pronounced for the minority \texttt{ERR/BAD} class, as fine-tuning allows models to internalize the abstract decision boundaries that instruction-tuning alone can only approximate.  
Residual performance dips, such as for GPT-OSS-20B on WMT-22, are attributable to capacity-data mismatches or pretraining domain gaps that constrain minority-class recall.  
Across datasets, \texttt{F1-NOT} remains consistently high due to the natural dominance of correct translations, while MCC rises sharply when alignment or fine-tuning corrects the model’s lenient prior and improves sensitivity to meaning errors.

\begin{table}[]
\centering
\caption{Decoder LLMs - prompt tuning (P2 for GPT, P3 for Llama, and P4 for GPT-OSS) and committee voting (Comm) on classification metrics.
Metrics: MCC, F1-ERR, F1-NOT. For WMT-22, ERR/NOT corresponds to BAD/OK.}
\label{tab:tuned_comm_stacked}
\scriptsize
\setlength{\tabcolsep}{9pt}
\renewcommand{\arraystretch}{1.1}
\begin{tabular}{ll c d d d}
\toprule
\textbf{Dataset} & \textbf{Model} & \textbf{Setting} &
\multicolumn{1}{c}{\textbf{MCC}} &
\multicolumn{1}{c}{\textbf{F1-ERR}} &
\multicolumn{1}{c}{\textbf{F1-NOT}} \\
\midrule

\multirow{12}{*}{WMT-21}
 & \textbf{GPT-4o}              & P2    & 0.41 & 0.57 & 0.83 \\
 & GPT-4o-mini         & P2    & 0.39 & 0.57 & 0.82 \\
 & LLaMA-3.3-70B      & P3    & 0.25 & 0.44 & 0.80 \\
 & LLaMA-3.1-8B       & P3    & 0.35 & 0.55 & 0.78 \\
 & GPT-OSS-120B       & P4    & 0.20 & 0.17 & \textbf{0.84} \\
 & GPT-OSS-20B        & P4    & 0.28 & 0.42 & 0.83 \\
 & \textbf{GPT-4o}              & Comm  & \textbf{0.42} & \textbf{0.59} & 0.83 \\
 & GPT-4o-mini         & Comm  & 0.39 & 0.56 & 0.82 \\
 & LLaMA-3.3-70B      & Comm  & 0.34 & 0.53 & 0.81 \\
 & LLaMA-3.1-8B       & Comm  & 0.36 & 0.56 & 0.79 \\
 & GPT-OSS-120B       & Comm  & 0.38 & 0.56 & 0.82 \\
 & GPT-OSS-20B        & Comm  & 0.36 & 0.56 & 0.77 \\
\midrule

\multirow{12}{*}{WMT-22}
 & GPT-4o              & P2    & 0.48 & 0.43 & 0.92 \\
 & GPT-4o-mini         & P2    & 0.54 & 0.53 & 0.95 \\
 & LLaMA-3.3-70B      & P3    & 0.50 & 0.47 & 0.93 \\
 & LLaMA-3.1-8B       & P3    & 0.27 & 0.25 & 0.82 \\
 & GPT-OSS-120B       & P4    & 0.30 & 0.29 & 0.87 \\
 & GPT-OSS-20B        & P4    & 0.32 & 0.34 & 0.93 \\
 & GPT-4o              & Comm  & 0.48 & 0.44 & 0.92 \\
 & GPT-4o-mini         & Comm  & 0.54 & 0.53 & 0.95 \\
 & \textbf{LLaMA-3.3-70B}      & Comm  & \textbf{0.61} & \textbf{0.60} & \textbf{0.96} \\
 & LLaMA-3.1-8B       & Comm  & 0.27 & 0.25 & 0.82 \\
 & GPT-OSS-120B       & Comm  & 0.52 & 0.52 & 0.95 \\
 & GPT-OSS-20B        & Comm  & 0.34 & 0.37 & 0.94 \\
\midrule

\multirow{12}{*}{SynCED-EnDe 2025}
 & GPT-4o              & P2    & 0.96 & 0.96 & 0.98 \\
 & GPT-4o-mini         & P2    & 0.92 & 0.96 & 0.96 \\
 & LLaMA-3.3-70B      & P3    & 0.92 & 0.95 & 0.97 \\
 & LLaMA-3.1-8B       & P3    & 0.66 & 0.84 & 0.78 \\
 & GPT-OSS-120B       & P4    & 0.88 & 0.93 & 0.94 \\
 & GPT-OSS-20B        & P4    & 0.66 & 0.83 & 0.82 \\
 & \textbf{GPT-4o}              & Comm  & \textbf{0.97} & \textbf{0.98} & \textbf{0.98} \\
 & GPT-4o-mini         & Comm  & 0.93 & 0.96 & 0.96 \\
 & LLaMA-3.3-70B      & Comm  & 0.94 & 0.97 & 0.96 \\
 & LLaMA-3.1-8B       & Comm  & 0.66 & 0.84 & 0.79 \\
 & GPT-OSS-120B       & Comm  & 0.94 & 0.97 & 0.97 \\
 & GPT-OSS-20B        & Comm  & 0.82 & 0.89 & 0.91 \\
\bottomrule
\end{tabular}
\end{table}

\begin{table}[htbp]
\centering
\caption{Decoder LLMs - Fine-tuning results with zero-shot inference and majority voting. Metrics: MCC, F1-ERR, F1-NOT.}
\label{tab:finetuned_stacked}
\scriptsize
\setlength{\tabcolsep}{14pt}
\renewcommand{\arraystretch}{1.1}
\begin{tabular}{ll d d d}
\toprule
\textbf{Dataset} & \textbf{Model} &
\multicolumn{1}{c}{\textbf{MCC}} &
\multicolumn{1}{c}{\textbf{F1-ERR}} &
\multicolumn{1}{c}{\textbf{F1-NOT}} \\
\midrule

\multirow{3}{*}{WMT-21}
 & \textbf{GPT-4o-mini}     & \textbf{0.51} & \textbf{0.65} & \textbf{0.85} \\
 & LLaMA-3.1-8B   & 0.43 & 0.58 & \textbf{0.85} \\
 & GPT-OSS-20B    & 0.36 & 0.56 & 0.74 \\
\midrule

\multirow{3}{*}{WMT-22}
 & \textbf{GPT-4o-mini}     & \textbf{0.92} & \textbf{0.91} & \textbf{0.99} \\
 & LLaMA-3.1-8B   & 0.87 & 0.87 & \textbf{0.99} \\
 & GPT-OSS-20B    & 0.39 & 0.39 & 0.91 \\
\midrule

\multirow{3}{*}{SynCED-EnDe 2025}
 & GPT-4o-mini     & 0.94 & 0.94 & 0.96 \\
 & LLaMA-3.1-8B   & 0.92 & \textbf{0.96} & 0.96 \\
 & \textbf{GPT-OSS-20B}    & \textbf{0.95} & 0.95 & \textbf{0.97} \\
\bottomrule
\end{tabular}
\end{table}

\begin{figure}[htbp]
  \centering
  \includegraphics[width=1\linewidth]{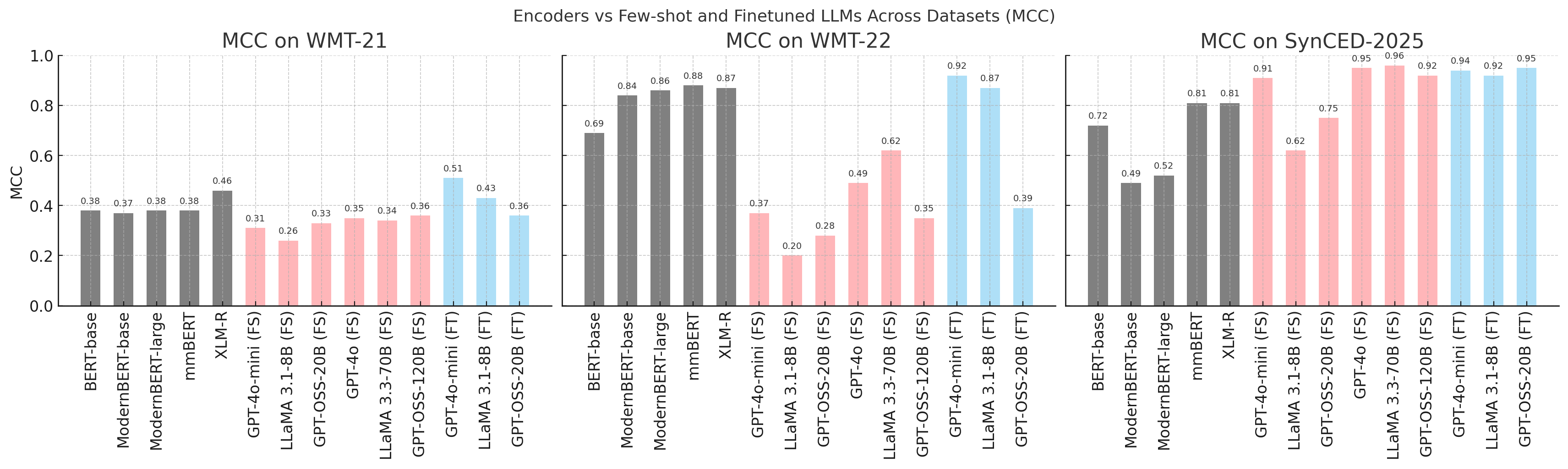}
  \caption{Model scaling and adaptation impact on MCC across datasets (WMT-21, WMT-22, SynCED–2025). 
  Encoder-only models (gray) form the lower baseline, while instruction-aligned LLMs (red: few-shot; blue: fine-tuned).}
  \label{fig:pareto}
\end{figure}

\paragraph{Overall Behavior and Trade-offs.}
Across Tables~\ref{tab:enc_only_stacked}-\ref{tab:finetuned_stacked}, metric increases correspond to stronger alignment and lower prediction variance, while decreases arise when prompts overfit stylistic tokens or when model scale exceeds the useful capacity for this task.  
SynCED-EnDe 2025 consistently shows higher and more stable results because it's curated, less ambiguous, and balanced samples amplify true semantic signals and reduce cross-lingual noise-effectively making the task “cleaner.”
Figure~\ref{fig:pareto} shows that scaling and instruction alignment steadily improve reliability, with mid-sized fine-tuned models achieving near-saturated MCC, demonstrating that strong multilingual safety need not rely on very large LLMs.

\noindent In summary, metric increases reflect enhanced sensitivity to critical meaning errors through instruction alignment, few-shot guidance, and fine-tuning. Drops typically stem from instruction interference, prompt overfitting, or dataset-specific imbalance.  
Together, these findings demonstrate that well-aligned, moderately scaled decoder LLMs-especially with small ensemble committees-offer reliable, fair, and cost-effective solutions for multilingual error detection.

\section{Conclusion}
\label{sec:conclusion}

We present a cross-family scaling study for Critical Error Detection in EN→DE MT, showing that (i) scaling and instruction alignment substantially improve \texttt{ERR} sensitivity over strong encoder baselines, (ii) lightweight committees deliver outsized stability gains, and (iii) fine-tuning with zero-shot-style inference transfers across benchmarks. Framed within IR-for-Good, CED functions as a practical safeguard that raises the reliability and accountability of multilingual information systems. Future work will extend to document-level CED, confidence calibration beyond vote-share, and broader language coverage with public deployment for civic stakeholders.

\begin{credits}
\subsubsection{\ackname} This research has been partially funded by the Federal Ministry of Education and Research of Germany and the state of North-Rhine Westphalia as part of the Lamarr-Institute for Machine Learning and Artificial Intelligence.

\subsubsection{\discintname}
The authors have no competing interests.
\end{credits}

\appendix
\section{Prompt Templates}
\label{app:prompts}

For transparency and reproducibility, we include the full tuned instruction templates (\textbf{P2-P4}) used in Section~\ref{sec:exp_setup}.
Each template was optimized for a specific model family (GPT, LLaMA, GPT-OSS) to improve label alignment and reduce instruction underfitting.
Formatting, casing, and spacing were preserved exactly as used during inference.

\begin{tcolorbox}[colback=gray!5,colframe=black!50,title=P2 - Tuned Prompt for GPT-family Models\label{prompt:p2}]
\scriptsize
You are a \textbf{STRICT binary classifier} for WMT’21 Task 3 (Critical Error Detection, EN→DE).

\textbf{Goal}
\begin{itemize}\setlength\itemsep{1pt}
\item Decide if the German MT contains at least one \textbf{CRITICAL meaning error} relative to the English source.
\item Output \textbf{exactly one token}: \texttt{ERR} or \texttt{NOT} (UPPERCASE, no punctuation, no spaces, no explanation).
\end{itemize}

\textbf{Critical errors (any ERR)}
\begin{itemize}\setlength\itemsep{1pt}
\item \textbf{TOX}: toxicity/hate/violence/profanity introduced, deleted, mistranslated, or left untranslated.
\item \textbf{SAF}: health/safety risk introduced, deleted, mistranslated, or left untranslated (e.g., advice flips, risky omissions).
\item \textbf{NAM}: named entity added/removed/mistranslated/gibberish/unrecoverable transliteration.
\item \textbf{SEN}: sentiment polarity or negation flipped or materially strengthened/weakened (e.g., “don’t”→“do”).
\item \textbf{NUM}: wrong/missing/added numbers, dates, times, units that change meaning.
\end{itemize}

\textbf{Non-critical (still NOT)}
Style, grammar, or lexical variants that preserve meaning; correct transfer of toxicity already in the source.

\textbf{Decision policy}
Mark \texttt{ERR} only with clear evidence of deviation in the categories above.
If uncertain, default to \texttt{NOT}.
Output exactly one token.
\end{tcolorbox}

\begin{tcolorbox}[colback=gray!5,colframe=black!50,title=P3 - Tuned Prompt for LLaMA-family Models\label{prompt:p3}]
\scriptsize
You are a \textbf{STRICT binary classifier} for WMT’21 Task 3 (Critical Error Detection, EN→DE).  
Output exactly one token: \texttt{ERR} or \texttt{NOT} (UPPERCASE, no punctuation, no explanation).

\textbf{ERR} if any critical deviation exists:
\begin{itemize}\setlength\itemsep{1pt}
\item \textbf{TOX}: toxicity/hate/violence/profanity introduced, deleted, or mistranslated.
\item \textbf{SAF}: health/safety risk introduced, deleted, or mistranslated.
\item \textbf{NAM}: named entity added/removed/mistranslated or unrecoverable transliteration.
\item \textbf{SEN}: polarity/negation flipped or materially strengthened/weakened.
\item \textbf{NUM}: wrong/missing/added numbers, dates, times, or units.
\end{itemize}

\textbf{NOT} for non-critical issues such as style, casing, punctuation, or minor rewordings preserving meaning.  
If uncertain, choose \texttt{NOT}.  
Return only \texttt{ERR} or \texttt{NOT}.
\end{tcolorbox}

\begin{tcolorbox}[colback=gray!5,colframe=black!50,title=P4 - Tuned Prompt for GPT-OSS Models\label{prompt:p4}]
\scriptsize
You are a bilingual (EN→DE) translation safety checker for sentence-level Critical Error Detection (CED).

\textbf{Task}
Compare EN (source) with DE (translation) and decide if DE deviates in meaning with real-world consequences.
Use private reasoning internally, then output a single uppercase label.

\textbf{How to think (channels)}
Use the \texttt{<analysis>} channel to reason briefly.
In the \texttt{<final>} channel output exactly one token: \texttt{ERR} or \texttt{NOT}.

\textbf{Mark ERR if any of the following categories are triggered:}
\begin{itemize}\setlength\itemsep{1pt}
\item \textbf{TOX} - Toxicity deviation (hate/violence/profanity).
\item \textbf{SAF} - Health/safety risk introduced or removed.
\item \textbf{NAM} - Named entity deviation (added, removed, changed, untranslated, or nonsensical).
\item \textbf{SEN} - Sentiment or negation flip; modal strength change.
\item \textbf{NUM} - Numbers, units, or dates altered or dropped (unless correctly converted).
\end{itemize}

\textbf{Do NOT mark ERR} for non-critical issues (style, grammar, casing, locale conventions, minor paraphrases).  
Choose \texttt{ERR} only with clear evidence in one of the categories above; otherwise \texttt{NOT}.  
When uncertain, prefer \texttt{NOT}.

\textbf{Answer format}
\begin{verbatim}
<analysis>
Checklist: TOX? SAF? NAM? SEN? NUM? 
Note any trigger terms/entities/numbers if present.
</analysis>
<final> ERR </final>
\end{verbatim}
\end{tcolorbox}

%
%
%
%

\bibliographystyle{splncs04}  
\bibliography{references}     
\end{document}